\documentclass[letterpaper, 10pt, conference]{ieeeconf}

\IEEEoverridecommandlockouts
\usepackage{amsmath,amssymb,amsfonts}
\usepackage{bbm}
\usepackage{graphicx}
\usepackage{booktabs}
\usepackage{multirow}
\usepackage[bookmarks=false,colorlinks=true,citecolor=blue,urlcolor=blue,linkcolor=black]{hyperref}
\usepackage{xcolor}
\usepackage{siunitx}
\graphicspath{{fig/}}
\title{\LARGE \bf
Visual Navigation Transformer with Pose Attention}

\author{Beiming Li$^{1}$, Jaime Romero$^{1}$, Jonathan Diller$^{1}$, Vijay Kumar$^{1}$, and Alejandro Ribeiro$^{2}$%
\thanks{$^{1}$B. Li, J. Romero, J. Diller, and V. Kumar are with the GRASP Laboratory, University of Pennsylvania, Philadelphia, PA 19104, USA
        {\tt\small \{beimingl, jaimerom, diller, kumar\}@\allowbreak engineering.upenn.edu}}%
\thanks{$^{2}$A. Ribeiro is with the Department of Electrical and Systems Engineering, University of Pennsylvania, Philadelphia, PA 19104, USA
        {\tt\small aribeiro@engineering.upenn.edu}}%
}

\begin{document}

\maketitle
\thispagestyle{empty}
\pagestyle{empty}

\begin{abstract}
Learned navigation policies typically consume observations as a temporally ordered history, with positional encodings tying each observation to when it was seen, making it difficult to reuse experience from earlier traversals of an environment. Systems that do reuse such experience usually construct an explicit representation, such as a map or a topological graph, and plan on it. We propose VNT-PA (Visual Navigation Transformer with Pose Attention), a transformer planner whose context is a set of depth keyframes indexed by camera pose. With camera poses as positional encoding, attention depends on the pose differences between keyframes rather than on their temporal order. VNT-PA is trained to imitate a shortest-path planner operating on the ground-truth scene mesh, predicting actions by querying the spatial context with only its current pose and the goal position. On point-goal navigation in HM3D validation scenes, VNT-PA reaches 93.3\% success and 90.4\% success weighted by path length (SPL), outperforming baselines that encode the same context as a temporal sequence or treat pose as an input feature, in both navigation performance and training efficiency. Because the spatial context is a pose-indexed set, frames from different trajectories can be fused at test time. The planner also degrades more gracefully under localization noise than a conventional baseline which plans on explicit maps. These results show that pose-stamped experience can serve directly as the environment representation for a learned planner, and that making attention depend on pose differences, rather than on temporal order, speeds up training and improves long-horizon navigation.

\end{abstract}

\section{Introduction}
\label{sec:intro}

Learned navigation policies, from end-to-end point-goal agents ~\cite{ddppo, poliformer} to navigation foundation models ~\cite{vint, nomad, navid, uninavid}, almost universally consume the robot's past observations as a temporal stream. Observations are kept as a sequence of time-indexed tokens~\cite{navid, uninavid, hamt}, truncated to a sliding window~\cite{vint, nomad}, or compressed into a recurrent state~\cite{ddppo}. However, such a memory has no natural position for observations from an earlier visit or from another robot, making it nontrivial to transfer earlier experience in an environment as context for a new task. Moreover, its cost grows with the length of the history, rather than with the size of the environment. Systems that do exploit long-horizon memory instead separate decision making from trajectory planning~\cite{remembr, raven}. They retain a persistent memory of past observations, but retrieve from it only to select navigation goals, while trajectory planning is offloaded to a conventional stack operating on a metric or occupancy map. 

In this work, we propose VNT-PA, a transformer planner that treats a pose-indexed set of keyframes as the environment representation, and plans near-optimal paths in previously explored environments by querying the context directly through pose-dependent attention, without the need for an explicit map. The set of encoded depth frames indexed by camera pose is referred to as the \textit{spatial context}, in contrast to the \textit{temporal context} of sequence models. The planner uses camera poses as rotary positional embeddings~\cite{roformer}, so that attention depends on pose differences between frames, rather than on their temporal order. At each decision, a query token stamped at the robot pose, knowing the goal but not current observation, reads the spatial context and predicts actions towards the goal.

We study the geometric core of the problem through point-goal navigation tasks, where a shortest-path expert provides exact supervision and every gain can be attributed to how the memory is organized and processed. Concretely, we make the following contributions:
\begin{itemize}\setlength{\itemsep}{1pt}
\item[\textbf{C1}] We formulate the spatial context, a pose-indexed set of keyframe features, as the environment representation. It is organized by where observations were taken rather than when, and is consumed directly by a learned planner, circumventing the need for an explicit map. A keyframe filter ties the context size to the space observed rather than to the length of the stream.
\item[\textbf{C2}] We introduce VNT-PA, a transformer planner trained to imitate a shortest-path expert which utilizes the ground-truth navigation mesh. It reads the spatial context using camera poses as rotary positional embedding, so that attention relates two entries only through the pose difference and is invariant to the temporal order of the entries and to translations of the world frame.
\item[\textbf{C3}] We show empirically, in HM3D validation scenes, that pose as positional embedding provides a beneficial inductive bias: VNT-PA handles long and detour-heavy episodes better than models that process the same keyframes as a temporal sequence, through a recurrent state, or with pose as an input feature.
\item[\textbf{C4}] We showcase desirable properties of our framework. First, the spatial context can be extended at test time with additional frames, such as previous frames discarded by the keyframe filter or the observations collected online, and VNT-PA benefits from them without retraining. Second, under localization error, our method degrades more gracefully than a conventional planner that relies on an occupancy map built from the same frames and poses that form our spatial context.
\end{itemize}

\section{Related Work}
\label{sec:related}
\textit{Explicit environment representations.} Most navigation systems fuse observations into a spatial data structure and plan on it. Occupancy grids and signed distance fields~\cite{hornung2013octomap, voxblox} integrate each observation into voxels by ray casting from the estimated sensor pose. Scene reconstructions from NeRF~\cite{nerf} and Gaussian Splatting~\cite{gaussiansplatting} are used directly as the map, with the planner querying the density field as collision cost~\cite{nerfnav} or extracting safe corridors from the Gaussians~\cite{splatnav}. Many learning-based systems still rely on an explicit map structure to store and organize observation features, even those with a learned planner trained end-to-end. Prior work predicts occupancy maps with a learned mapper~\cite{ans}, writes vision features to a 2D grid~\cite{cmp, neuralmap}, or stores vision-language features in a map~\cite{vlmaps, dynam3d}. All these approaches ultimately represent the environment using a predefined map structure, constrained by choices such as resolution, spatial extent, and topology.

\textit{Time-indexed memory.} Learned navigation policies that retain past observations typically organize memory according to the order in which those observations were acquired. Video-based navigators tokenize frames in order of arrival and aggregate them by recency~\cite{navid, uninavid}, while end-to-end point-goal navigators rely on recurrent state~\cite{ddppo}. Image-goal policies attend over a few recent frames and defer long-horizon memory to a separate topological graph~\cite{vint, nomad}. Spatially-Enhanced Recurrent Units (SRU)~\cite{sru} address the weak spatial memorization of recurrent networks with a gate that modulates the state by a transformation of the input, but they still integrate observations in order of arrival.

\textit{Memory-augmented agents.} Recent systems built on foundation models retrieve from a persistent memory of past observations to answer queries and select navigation goals~\cite{remembr, raven, mobilityvla}. However, they maintain two representations of the same environment. The foundation model reasons over stored observations to select a navigation goal, while path planning relies on a conventional stack with a separately maintained metric or topological map.

\textit{Geometry-aware attention.} A separate line of work investigates how geometric structure can enter attention directly. RoPE~\cite{roformer} rotates queries and keys by their absolute positions so that the attention logit depends only on their relative position. GTA~\cite{gta} and PRoPE~\cite{prope} extend this idea to multi-view transformers by incorporating relative camera transforms in attention, and show advantages over supplying camera pose as a content feature~\cite{srt}.

Navigation systems above reuse past observations in one of two ways. One family converts observations to an explicit spatial representation before planning. The other retains observations directly but organizes them according to \emph{when} they were acquired. We pursue a third alternative: organizing memory by \emph{where} the observations were acquired. Closest to our approach is the Scene Memory Transformer~\cite{smt}, which attends over embedded observations as an unordered set but supplies pose as a token feature. We use pose as the positional embedding instead, and show that this choice improves long-horizon navigation and training efficiency.

\section{Visual Navigation Transformer \\with Pose Attention}
\label{sec:method}

We introduce VNT-PA, a transformer planner that navigates to point goals using prior experience of an environment, represented as a pose-indexed set of depth keyframes. We introduce the problem setting in Sec.~\ref{sec:setting}, followed by the spatial context and its positional embedding in Sec.~\ref{sec:attention}, keyframe encoding and filtering in Sec.~\ref{sec:context}, model architecture in Sec.~\ref{sec:arch}, and training setup in Sec.~\ref{sec:training}.

\begin{figure*}[t]
\centering
\includegraphics[width=0.98\textwidth]{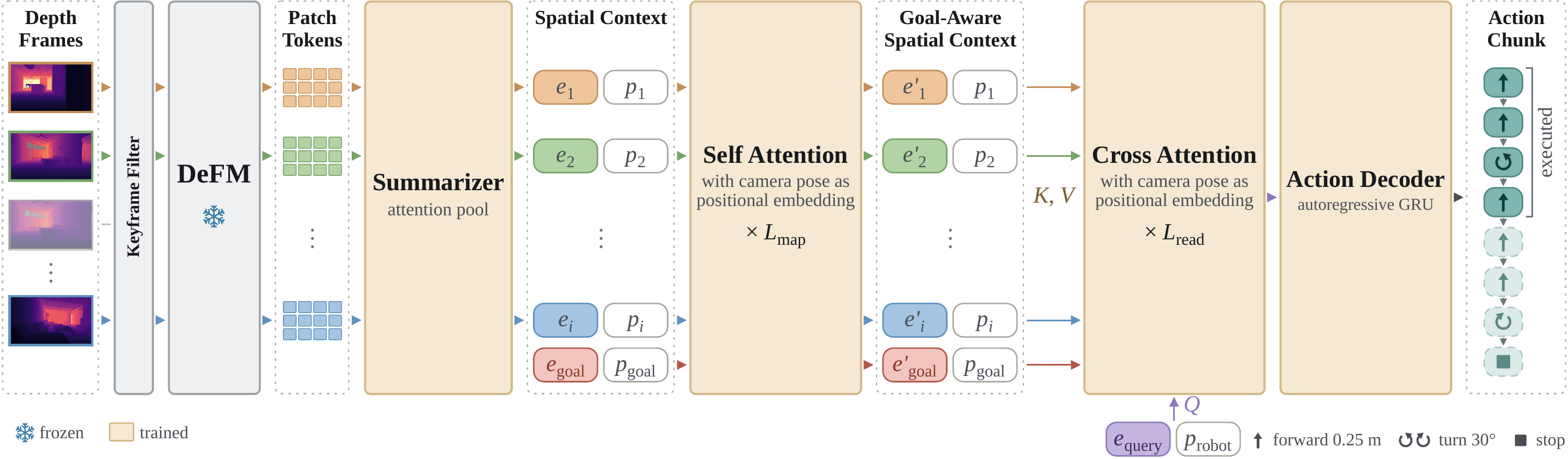}
\caption{Overview of VNT-PA. (1) A keyframe filter keeps only depth frames that observe sufficient new space. A frozen depth encoder (DeFM) and a learned attention-pooling summarizer represent each keyframe as a single token $e_i$. (2) The set formed by image tokens indexed by corresponding camera poses is referred to as spatial context. It serves as an implicit representation of the environment. Together with a learned goal token $e_{\mathrm{goal}}$ stamped at the goal position $p_{\mathrm{goal}}$, the context set passes through $L_{\mathrm{map}}$ bidirectional self-attention blocks that use poses as rotary positional embeddings. The output embedding is referred to as goal-aware spatial context. (3) A query token $e_{\mathrm{query}}$ is built from the relative goal location and stamped at the robot pose $p_{\mathrm{robot}}$. It cross-attends the goal-aware context through $L_{\mathrm{read}}$ cross-attention blocks with the same positional embedding. (4) An autoregressive GRU decodes the final query token embedding into a chunk of actions. The robot executes the first $h$ actions and then queries the planner from its new pose.}
\label{fig:arch}
\end{figure*}

\subsection{Problem Setting}
\label{sec:setting}

We consider a mobile robot initialized at a random pose in an indoor environment and tasked with reaching a goal position. The setting resembles PointGoal navigation~\cite{ddppo}: the robot executes the discrete actions \textsc{forward}, \textsc{turn-left}, \textsc{turn-right}, and \textsc{stop}, and it succeeds if it calls \textsc{stop} within a geodesic distance $r$ of the goal within $T$ steps. We assume that the robot knows its planar pose $p_{\mathrm{robot}}=(x_{\mathrm{robot}},y_{\mathrm{robot}},\psi_{\mathrm{robot}})$ and the goal position $p_{\mathrm{goal}}=(x_{\mathrm{goal}},y_{\mathrm{goal}})$ in a common world frame, whose origin can be chosen arbitrarily.

The main difference from standard PointGoal navigation is that the robot is given the depth frames $I_1,\dots,I_N$ recorded during earlier traversals, for example an exploration run, together with their camera poses $p_1,\dots,p_N$ in the same world frame. The planner acts only on these earlier frames, the robot pose, and the goal, without relying on the current view or any frame recorded during the ongoing episode. Thus, the earlier frames must serve as the environment representation. In practice, it suffices to localize the start pose once relative to these depth frames and track the robot pose during navigation, for example, with LiDAR-inertial odometry~\cite{xu2022fast}. This setting resembles real-world deployments in which a robot operates in the same environment over time and has a long stream of observations.

\subsection{Pose-Indexed Spatial Context}
\label{sec:attention}

\textbf{Spatial context.} A transformer relates the tokens of its context through attention, and the positional embedding determines which relations attention can express. Sequence models index each observation by its time of capture~\cite{navid,chen2021decision}. However, the temporal indices of observations captured in the same environment across different times or trajectories are arbitrary. To represent an environment, the natural index of an observation is instead the pose from which it was captured. We therefore index each frame by its camera pose and the context becomes an unordered set of pose-stamped entries,
\begin{equation}
\mathcal{S}=\big\{(e_i,p_i)\big\},
\label{eq:context}
\end{equation}
where $e_i\in\mathbb{R}^{d}$ is a feature vector computed from frame $I_i$ and $p_i=(x_i,y_i,\psi_i)$ is its camera pose. We refer to $\mathcal{S}$ as the \emph{spatial context}, in contrast to the \emph{temporal context} used by sequence models.

\textbf{Rotary position embedding.} RoPE~\cite{roformer} makes attention depend on the relative positions of tokens. Consider an attention head whose query and key vectors have dimension $2n$, and a token at a scalar position $t$, such as its index in an ordered sequence. RoPE divides the $2n$-dimensional space into $n$ two-dimensional subspaces and rotates the \mbox{$m$-th} subspace by the angle $\theta_m t$,
\begin{equation}
\mathbf{R}_{\theta}(t)=\bigoplus_{m=1}^{n}\rho(\theta_m t),\qquad \theta_m=b^{-(m-1)/n},
\label{eq:rope1d}
\end{equation}
where $\rho(\alpha)$ denotes the $2\times2$ rotation matrix of angle $\alpha$, $\bigoplus$ stacks its arguments into a block-diagonal matrix, and $b$ is the base of the geometric frequencies $\theta=(\theta_1,\dots,\theta_n)$. Queries and keys are multiplied by $\mathbf{R}_{\theta}$ at their own positions, so that the attention logit depends on the positions only through their difference, as Eq.~\eqref{eq:relative} shows for poses. The temporal baselines in Sec.~\ref{sec:exp} use this embedding, with $t$ denoting the temporal index of each observation in its sequence. We refer to this as temporal RoPE.

\textbf{Pose RoPE.} Entries of the spatial context are indexed by camera pose. Since the origin of the world frame is arbitrary, attention should depend on poses only through their differences. We therefore apply the rotary embedding of Eq.~\eqref{eq:rope1d} to each pose component. A head of dimension $6n$ is divided into $3n$ two-dimensional subspaces, with $n$ subspaces assigned to each component. The overall rotation matrix is:
\begin{equation}
\mathbf{R}(p)=\mathbf{R}_{\omega}(x)\oplus\mathbf{R}_{\omega}(y)\oplus\mathbf{R}_{\nu}(\psi),
\label{eq:rope}
\end{equation}
where $\omega$ and $\nu$ denote the frequencies of the position and heading components. Since $\mathbf{R}(p_i)^{\top}\mathbf{R}(p_j)=\mathbf{R}(p_j-p_i)$, the attention logit between a query at pose $p_i$ and a key at pose $p_j$ becomes
\begin{equation}
\big\langle\mathbf{R}(p_i)\,\mathbf{q}_i,\ \mathbf{R}(p_j)\,\mathbf{k}_j\big\rangle=\mathbf{q}_i^{\top}\,\mathbf{R}(p_j-p_i)\,\mathbf{k}_j,
\label{eq:relative}
\end{equation}
where $p_j-p_i=(x_j-x_i,\,y_j-y_i,\,\psi_j-\psi_i)$. The logit is thus invariant to translations of the world frame.

\textbf{Choice of frequencies.} Since planar positions $x$ and $y$ are metric, we scale the geometric frequencies by a scalar $s$, which gives $\omega_m=s^{-1}b^{-(m-1)/n}$. The highest frequency advances one radian per $s$ meters. A smaller $s$ thus separates nearby poses more sharply but is more sensitive to localization error. The lowest frequency is $\omega_n=s^{-1}b^{-(n-1)/n}$, which corresponds to the longest wavelength $\lambda_{\max}=2\pi s\,b^{(n-1)/n}$. For a displacement longer than $\lambda_{\max}/2$, the rotation angle of the lowest frequency exceeds $\pi$ and wraps around, producing the same rotation as a shorter displacement in the opposite direction. Therefore, we choose $b$ such that $\lambda_{\max}/2$ exceeds the extent of most environments along each axis. 

Unlike positions, headings are periodic, so $\mathbf{R}_{\nu}(\psi+2\pi)=\mathbf{R}_{\nu}(\psi)$ must hold, which is the case if every $\nu_m$ is an integer. We choose $\nu_m=m$, so that the heading part of each attention logit becomes a Fourier series in $\psi_j-\psi_i$ up to the $n$-th harmonic, with coefficients defined by queries and keys. The first harmonic distinguishes all heading differences, and higher harmonics resolve finer ones.

\textbf{Rotation augmentation.} Our RoPE expresses the distance between two tokens in the world frame rather than in the ego frame of the query. Attention is therefore invariant to translations of the world frame but not to its rotations. Rather than building rotation invariance into the architecture, we let the model learn it from data with rotation augmentation. During training, the world frame of each example is rotated by an angle $\gamma\sim\mathcal{U}[0,2\pi)$, which changes the poses of the keyframes, the robot, and the goal.

\subsection{Frame Embedding and Filtering}
\label{sec:context}

\textbf{Frame embedding.} Each frame in the spatial context is encoded by a frozen depth foundation model, DeFM~\cite{defm}, into $G\times G$ patch tokens. Then, a learned attention-pooling summarizer compresses the patch tokens into a single feature vector $e_i\in\mathbb{R}^{d}$. We use depth images rather than RGB images because the decisions the planner learns depend solely on the geometry of the environment.

\textbf{Keyframe filter.} Since the cost of attention grows with the number of entries, and depth frames captured at nearby poses overlap heavily, we keep only the frames that discover sufficient amount of new information. Spatial context facilitates such filtering since removing an entry does not disturb the index of any other. 

We approximate the space a frame observes by dividing it into $G\times G$ patches, unprojecting the median depth of each patch along its central ray with the camera intrinsics and pose, and quantizing the point to a voxel tagged with its viewing direction, so that a surface seen from substantially different directions counts as distinct. Let $\mathcal{C}(I_i,p_i)$ denote the resulting voxel set. Frames are processed sequentially, and frame $i$ joins the set of keyframes $\mathcal{K}$ if it observes at least $\eta$ voxels that no earlier keyframe observed,
\begin{equation}
i\in\mathcal{K}\ \iff\ \Big|\,\mathcal{C}(I_i,p_i)\setminus\textstyle\bigcup_{j\in\mathcal{K},\,j<i}\mathcal{C}(I_j,p_j)\,\Big|\ \ge\ \eta.
\label{eq:gate}
\end{equation}

Keyframes construct a lightweight spatial context, $\mathcal{S}=\{(e_i,p_i): i\in\mathcal{K}\}$, and the cost of a decision is now tied to the size of the environment rather than to the length of the robot's history. 

\textbf{Extension.} Conversely, any depth frame with a pose in the same world frame can be added to $\mathcal{S}$, regardless of when or along which trajectory it was captured. For example, a robot navigating with a spatial context constructed from earlier travels can append its current view to $\mathcal{S}$ at each query. Sec.~\ref{sec:extension} evaluates such test-time extensions.

\subsection{Model Architecture}
\label{sec:arch}

\textbf{Goal-aware context.} The spatial context $\mathcal{S}$, together with a learned goal token $e_{\mathrm{goal}}$ stamped at the goal position $p_{\mathrm{goal}}$, is first processed by $L_{\mathrm{map}}$ bidirectional self-attention blocks with the positional embedding defined in Eq.~\eqref{eq:rope}. The goal has no heading, so the heading subspaces of its query and key vectors are set to zero. Every entry can thus relate itself to every other frame and to the goal through their pose differences before any decision is made. We call the output of self-attention blocks the goal-aware context.

\textbf{Query.} At each decision, the planner constructs a query token. Let $\Delta p=(\Delta x,\Delta y)$ denote the goal position expressed in the robot frame, and $\alpha=\operatorname{atan2}(\Delta y,\Delta x)$ its angle relative to the robot heading. The query token is $e_{\mathrm{query}}=f\big(\lVert\Delta p\rVert,\ \cos\alpha,\ \sin\alpha\big)$, where $f$ is a learned embedding. The query is stamped at the robot pose $p_{\mathrm{robot}}$ and cross-attends to the goal-aware context through $L_{\mathrm{read}}$ blocks with the same positional embedding. The query does not contain the robot's current or past observations. Therefore, VNT-PA resembles a conventional map-based planner in that it relies only on a representation of the environment, here the spatial context, along with robot pose and the goal. This design prevents the model from relying on the current view and ensures that each decision is grounded in the spatial context.

\textbf{Action decoder.} Let $\mathbf{z}$ denote the output embedding of the query token. A gated recurrent unit (GRU)~\cite{cho2014learning} decodes $\mathbf{z}$ into a chunk of $H$ actions autoregressively,
\begin{equation}
\mathbf{g}_{\ell}=\mathrm{GRU}\big(\mathbf{g}_{\ell-1},\,a_{\ell-1}\big),\qquad \mathbf{g}_0=\mathbf{z},
\label{eq:decoder}
\end{equation}
where $a_0$ is a learned start token. A linear layer maps $\mathbf{g}_{\ell}$ to one logit per action, whose softmax gives $\pi_{\theta}(a_{\ell}\mid\mathbf{z},a_{1:\ell-1})$. During training, the decoder is conditioned on expert actions. During execution, it is conditioned on the actual action history instead. The robot then executes the first $h\le H$ actions of the chunk and queries the planner again from its new pose. A smaller $h$ replans more often, while a larger $h$ commits to longer and potentially more consistent motion.

\subsection{Training}
\label{sec:training}
We train the planner by imitating an expert trajectory planner that utilizes the ground-truth mesh of scenes~\cite{habitat}. The planner is thus encouraged to learn geometric reasoning based on the spatial context.

Each training example consists of the spatial context built from depth frames of one environment and a randomly sampled navigation task, defined by a start pose $p_{\mathrm{robot}}$ and a goal position $p_{\mathrm{goal}}$. The planner is trained to minimize the following cross-entropy loss:
\begin{equation}
\mathcal{L}(\theta)=-\mathbb{E}\Big[\sum_{\ell=1}^{H}\log\pi_{\theta}\big(a^{*}_{\ell}\,\big|\,\tilde{\mathcal{S}},\tilde{p}_{\mathrm{robot}},p_{\mathrm{goal}},a^{*}_{1:\ell-1}\big)\Big],
\label{eq:loss}
\end{equation}
where $a^{*}_{1:\ell}$ are the first $\ell$ expert actions, and $\tilde{p}_{\mathrm{robot}}$ and $\tilde{\mathcal{S}}$ denote the noise-perturbed robot pose and spatial context.

\begin{table*}[t]
\centering
\caption{Closed-loop navigation performance with ground truth localization. Mean $\pm$ standard deviation over three training seeds.}
\label{tab:main}
\providecommand{\std}[1]{{\tiny$\pm$#1}}
\setlength{\tabcolsep}{2.5pt}
\resizebox{\textwidth}{!}{
\begin{tabular}{l cc cccccc cccccc}
\toprule
    & & & \multicolumn{6}{c}{Detour ratio $\kappa$} & \multicolumn{6}{c}{Geodesic distance (m)} \\
\cmidrule(lr){4-9} \cmidrule(lr){10-15}
    & \multicolumn{2}{c}{Overall} & \multicolumn{2}{c}{$<1.2$} & \multicolumn{2}{c}{$[1.2,1.5)$} & \multicolumn{2}{c}{$\ge1.5$} & \multicolumn{2}{c}{$<10$} & \multicolumn{2}{c}{$[10,15)$} & \multicolumn{2}{c}{$\ge15$} \\
\cmidrule(lr){2-3} \cmidrule(lr){4-5} \cmidrule(lr){6-7} \cmidrule(lr){8-9} \cmidrule(lr){10-11} \cmidrule(lr){12-13} \cmidrule(lr){14-15}
Method & SR & SPL & SR & SPL & SR & SPL & SR & SPL & SR & SPL & SR & SPL & SR & SPL \\
\midrule
LSTM~\cite{hochreiter1997long} & 31.9\std{0.6} & 28.5\std{0.6} & 43.1\std{1.3} & 38.6\std{1.1} & 29.3\std{0.7} & 26.2\std{0.7} & 11.3\std{0.8} & 9.6\std{0.7} & 40.3\std{0.7} & 35.9\std{0.6} & 15.8\std{0.6} & 14.4\std{0.6} & 5.7\std{0.3} & 5.1\std{0.3} \\
SRU~\cite{sru} & 30.3\std{1.5} & 27.3\std{1.4} & 43.0\std{1.6} & 38.9\std{1.3} & 24.8\std{1.2} & 22.3\std{1.3} & 11.4\std{1.8} & 10.0\std{1.6} & 38.2\std{1.5} & 34.4\std{1.4} & 14.4\std{1.8} & 13.1\std{1.8} & 7.8\std{1.1} & 7.1\std{1.1} \\
Causal Transformer & 90.3\std{0.5} & 87.3\std{0.5} & 94.1\std{0.7} & 91.0\std{0.5} & 88.7\std{0.8} & 85.9\std{0.8} & 84.5\std{0.8} & 81.2\std{0.8} & 92.8\std{0.5} & 89.8\std{0.5} & 86.4\std{0.7} & 83.2\std{0.6} & 80.4\std{1.7} & 77.4\std{1.3} \\
VNT-TA & 87.1\std{3.5} & 83.9\std{3.8} & 92.0\std{2.0} & 88.7\std{2.2} & 85.9\std{4.6} & 82.7\std{5.0} & 78.3\std{5.0} & 75.2\std{5.3} & 91.0\std{3.0} & 87.8\std{3.4} & 82.3\std{4.6} & 79.2\std{4.7} & 67.7\std{4.3} & 64.9\std{4.4} \\
VNT-PF & 89.5\std{0.3} & 86.5\std{0.4} & 94.0\std{0.4} & 90.9\std{0.4} & 88.3\std{1.0} & 85.4\std{0.9} & 81.5\std{0.7} & 78.5\std{0.8} & 92.9\std{0.6} & 89.8\std{0.6} & 85.0\std{1.6} & 82.2\std{1.7} & 73.6\std{1.5} & 70.8\std{1.3} \\
\midrule
VNT-PA (poses only) & 66.4\std{0.8} & 62.8\std{0.8} & 77.6\std{0.3} & 73.2\std{0.1} & 63.3\std{1.0} & 60.0\std{1.1} & 46.9\std{4.3} & 44.3\std{4.2} & 73.5\std{0.9} & 69.5\std{0.8} & 53.8\std{1.0} & 51.0\std{1.2} & 41.7\std{0.9} & 39.5\std{1.0} \\
VNT-PA (ours) & \textbf{93.3}\std{1.2} & \textbf{90.4}\std{1.1} & \textbf{96.1}\std{0.7} & \textbf{93.2}\std{0.6} & \textbf{92.9}\std{1.3} & \textbf{90.1}\std{1.1} & \textbf{87.5}\std{2.3} & \textbf{84.6}\std{2.3} & \textbf{95.2}\std{0.8} & \textbf{92.4}\std{0.7} & \textbf{90.2}\std{2.3} & \textbf{87.4}\std{2.1} & \textbf{85.4}\std{2.0} & \textbf{82.0}\std{1.9} \\
\bottomrule
\end{tabular}
}
\end{table*}

\begin{figure*}[htbp]
\centering
\includegraphics[width=\textwidth]{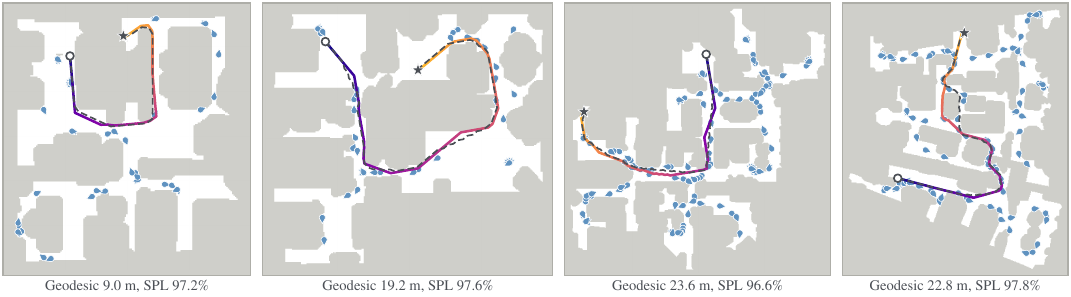}
\caption{Closed-loop navigation of VNT-PA in unseen evaluation scenes. White marks the navigable space of the ground-truth navigation mesh, and blue markers show the keyframe poses in the spatial context. The trajectory of VNT-PA is colored from dark at the start (circle) to bright at the goal (star), and the dashed line shows the expert trajectory.}
\label{fig:qualitative}
\end{figure*}

\section{Experiments}
\label{sec:exp}

Our experiments ask three questions. Can a learned planner navigate effectively using only a pose-indexed set of keyframes as its environment representation? Does using pose as positional embedding provide a useful inductive bias? And does representing the environment as a set offer practical benefits? We answer them by evaluating VNT-PA in closed loop against an expert planner operating on the ground-truth mesh, and baselines that process the same memory in different ways. We further test how the planner performs when its context is extended at test time or perturbed by localization noise. Finally, we investigate what information the model uses for planning by visualizing the attention pattern of the token summarizer.

\subsection{Experimental Setup}
\label{sec:setup}

\textbf{Simulation environments.} All experiments run in Habitat-Sim~\cite{habitat} with Habitat-Matterport 3D dataset (HM3D)~\cite{ramakrishnan2021habitat}, which contains 3D scans of real world indoor environments. We use 800 scenes for training and 100 scenes for validation. The simulated robot has a radius of 0.2\,m, and its depth camera is mounted 0.6\,m above the ground, with a resolution of $160\times272$ pixels and a horizontal field of view of $110^\circ$. \textsc{forward} action moves the robot forward by 0.25\,m, \textsc{turn-left} and \textsc{turn-right} rotate it by $30^\circ$. An episode succeeds if the robot calls \textsc{stop} within a geodesic distance of $r=0.2$\,m from the goal, within $T=500$ steps.

\textbf{Construction of spatial context.} We build the spatial context from depth frames observed along classic frontier-based exploration trajectories~\cite{yamauchi}. For each scene, we run explorations from five random starting poses. Episodes with short paths are discarded as they are collected in small buildings where navigation can be trivial. This process yields 3,834 bags of depth frames for training and 476 for evaluation. On average, an exploration contains 313 frames, covers a path of 54\,m, and observes 93\% of the area.

Each bag of frames is then individually passed through the keyframe filter with $G=16$ patches per side and a threshold of $\eta=10$. The filter keeps 21\% of the depth frames, yielding 67 keyframes per trajectory on average and at most 838.

\textbf{Architecture.} We use the frozen DeFM ViT-S/14 backbone~\cite{defm} to encode resized depth frames ($224\times224$) into $16\times16$ patch tokens of 384 dimension. The summarizer keeps this feature size. The planner has hidden dimension $d=768$, $L_{\mathrm{map}}=6$ self-attention blocks, $L_{\mathrm{read}}=4$ cross-attention blocks, each with 8 attention heads. It predicts a chunk of $H=8$ actions, executes $h=4$ of them per query. The model has 74.3M trainable parameters in total. 

Each attention head is of dimension 96, which gives $n=16$ subspaces per pose component. We set the scale of the pose RoPE to $s=0.25$\,m and the frequency base to $b=100$. The longest wavelength is then about 120\,m, more than twice the extent of most training environments along each axis.

On one NVIDIA L40S GPU, DeFM encodes a depth frame in 3.6\,ms, which is done once per frame. The planner is designed to be lightweight, it costs only 14\,ms for one inference step given 300 context frames.

\textbf{Training.} Expert trajectories are geodesic shortest paths to goal returned by Habitat-Sim’s built-in optimal path planner, which operates on the ground-truth environment mesh. Each trajectory is then converted into a sequence of discrete actions that serves as the prediction target.

Each training step samples 3,000 tuples of scene, start pose, and goal point. Let $\kappa$ denote the ratio of the geodesic distance between start and goal to their straight-line distance. A sample with $\kappa<1.1$ is rejected with probability \mbox{$1-20(\kappa-0.98)^2$} to downweight nearly straight episodes. The poses of the spatial context and of the robot are perturbed by uniform noise of up to $\pm0.1$\,m per axis and $\pm0.2$\,rad to simulate localization error.

All models are trained for 30,000 steps using AdamW, with $\beta_1=0.9$, $\beta_2=0.95$, weight decay of 0.05. The learning rate is warmed up to $3\times10^{-4}$ over the first 1,000 steps, and linearly decays to 5\% of its peak over the last 15\% of the steps. Training our planner takes approximately 4 hours on one NVIDIA L40S GPU.

\textbf{Evaluation.} We evaluate all models on a fixed set of 2,220 episodes. Each episode consists of a pre-collected exploration trajectory in a validation scene, a start pose, and a goal position. All evaluations are closed-loop: the robot executes the first $h$ actions of each decoded chunk in the simulator, and the planner is queried again from the resulting pose.

\subsection{Navigation Performance}
\label{sec:main}

We report the success rate (SR) and success weighted by path length (SPL) on validation scenes. SPL is $\sigma_i D_i/\max(P_i, D_i)$ averaged over all episodes, where $\sigma_i\in\{0,1\}$ indicates success, $D_i$ is the geodesic distance from start to goal, and $P_i$ is the length of our executed path. We compare our architecture with baselines that process the same context differently.

\textbf{Baselines.} Unlike sequential policies such as~\cite{chen2021decision}, all models in our comparison are structured like a map-based planner. At each decision step, a query constructed from the robot pose and the goal reads only the encoded keyframes, observations from the current episode are not included in the context.

All baselines share the frozen depth encoder, summarizer, and action decoder with VNT-PA, and have roughly the same number of trainable parameters. They treat the pose of every keyframe and the goal position as input features. We compare against (1) \textit{VNT-TA} (temporal attention): our architecture with the pose RoPE replaced by the temporal RoPE of Eq.~\eqref{eq:rope1d}, which indexes keyframes by their order in the exploration trajectory; (2) \textit{VNT-PF} (pose as feature): our architecture with no positional embedding and treats pose as an input feature, resembles Scene Memory Transformer~\cite{smt}; (3) \textit{Causal Transformer}: a standard decoder-only transformer with temporal RoPE, in which each entry attends only to earlier keyframes and the query is appended as the last token; (4) \textit{LSTM}~\cite{hochreiter1997long} and (5) \textit{SRU}~\cite{sru}: recurrent networks that integrate keyframes into a fixed-size hidden state. SRU extends LSTM by modulating the hidden state with a learned linear transformation of the input, which was proposed for long-range mapless navigation. In addition, \textit{VNT-PA (poses only)}: our architecture with every feature $e_i$ set to zero, so that the planner sees only keyframe poses and the goal. Each model is trained with three random seeds.

\textbf{Results.} Table~\ref{tab:main} reports SR and SPL in percentage as mean and standard deviation over three training seeds, for all episodes, also grouped by the detour ratio $\kappa$ and by the geodesic distance between start and goal. 

VNT-PA achieves the highest SR and SPL overall and in every group, with a large advantage on hard episodes. As the geodesic distance grows, the SPL gap between VNT-PA and VNT-TA widens from 4.6 to 17.1 points. Since long episodes require relating keyframes that are far apart, the widening gap suggests that these relations are easier to exploit when keyframes are indexed by pose rather than by time. VNT-TA shares our architecture and differs only in its positional embedding, yet its overall SPL drops by 6.5 points and falls below that of the Causal Transformer. VNT-PF removes the temporal order and recovers part of this loss, but it remains 3.8 points of success below VNT-PA overall and 11.8 points on episodes of at least 15\,m. The advantage of VNT-PA therefore stems from the pose RoPE rather than from the architecture itself. In contrast, both recurrent baselines succeed in only about 30\% of all episodes and in about 11\% of those with $\kappa\ge1.5$, suggesting that a fixed-size recurrent state struggles to serve as a representation of the environment. 

Without depth features, VNT-PA (poses only) succeeds in 66.4\% of all episodes. Although every keyframe pose marks a navigable position and is therefore a strong cue, the poses are far from sufficient. The performance beyond them comes from the geometry that the keyframes observed, which enters the planner only through their depth features.

Fig.~\ref{fig:qualitative} shows example trajectories of VNT-PA. Since the expert plans on the ground-truth mesh, matching its decisions in validation scenes from keyframes alone shows that the spatial context is a sufficient representation for planning.

\subsection{Extending Spatial Context}
\label{sec:extension}

If the context is treated as a temporally ordered sequence, frames observed at different times or from different sources have no natural position index. In contrast, any depth frame stamped with its camera pose can be used to extend the spatial context, either to cover previously unobserved areas or to add detail to already observed regions.

We extend the context of VNT-PA from two sources at test time. First, we add depth frames discarded by the keyframe filter, allowing the planner to access all frames from the exploration trajectory. This increases the context from an average of 67 to 313 entries per episode. Second, we add the robot's current depth view each time the planner is queried during execution. Table~\ref{tab:context} shows that both extensions improve the success rate, even though the planner is trained only on keyframes of exploration trajectories. For episodes of at least 15\,m, the success rate rises from 87.7\% to as much as 96.3\%. These results show that VNT-PA generalizes without retraining to spatial contexts that are either denser or include views from disjoint navigation trajectories, and that it benefits from these additional views.

\begin{table}[t]
\centering
\caption{Closed-loop navigation of VNT-PA (single seed) with its spatial context extended at test time.}
\label{tab:context}
\providecommand{\twolinehead}[2]{\begin{tabular}[b]{@{}c@{}}#1\\#2\end{tabular}}
\setlength{\tabcolsep}{4pt}
\small
\begin{tabular}{lc cc ccc}
\toprule
\multicolumn{2}{c}{Spatial context} & \multicolumn{2}{c}{Overall} & \multicolumn{3}{c}{SR by geodesic distance (m)} \\
\cmidrule(lr){1-2} \cmidrule(lr){3-4} \cmidrule(lr){5-7}
\twolinehead{Exploration}{frames} & \twolinehead{Robot}{views} & SR & SPL & $<10$ & $[10,15)$ & $\ge15$ \\
\midrule
Keyframes  &            & 94.6 & 91.7 & 96.2 & 92.6 & 87.7 \\
Keyframes  & \checkmark & 96.5 & 93.5 & 97.5 & 94.0 & 95.2 \\
All frames &            & 96.2 & 93.1 & 97.4 & 94.4 & 90.9 \\
All frames & \checkmark & \textbf{97.1} & \textbf{94.0} & \textbf{98.0} & \textbf{94.6} & \textbf{96.3} \\
\bottomrule
\end{tabular}
\end{table}

\subsection{Training Efficiency}

\begin{figure}[b]
\centering
\includegraphics[width=0.95\columnwidth]{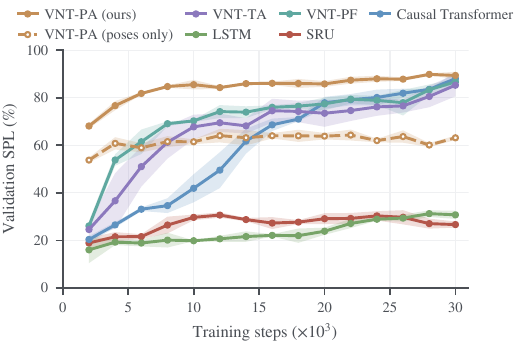}
\caption{Closed-loop SPL of intermediate checkpoints, mean $\pm$ standard deviation over models trained with three random seeds.}
\label{fig:efficiency}
\end{figure}

We evaluate training efficiency by measuring the SPL of intermediate checkpoints. As shown in Fig.~\ref{fig:efficiency}, VNT-PA reaches 80\% SPL within 5,000 steps across all three seeds, while the Causal Transformer, VNT-PF, and VNT-TA require more than 20,000 steps to reach the same performance. Temporal variants narrow this gap by the end of training, as they have access to poses as input features and also benefit from temporal adjacency as a proxy for spatial connectivity along the continuous exploration trajectory. Nevertheless, VNT-PA converges substantially faster, suggesting that pose RoPE provides a beneficial inductive bias.

\subsection{Robustness to Localization Error}
\label{sec:noise}

We compare VNT-PA against a conventional map-and-plan pipeline, labeled OctoMap + A*. It fuses depth frames into a probabilistic occupancy map~\cite{hornung2013octomap}, with a voxel edge of 0.05\,m. Every valid depth pixel casts a ray from the sensor pose, updating the voxels it traverses with a miss probability of $0.4$ and the voxel at its endpoint with a hit probability of $0.7$. The 3D occupancy map is projected onto a 2D grid, where obstacles are inflated by the robot radius, and the resulting grid is used for A* planning~\cite{hart1968formal}. We perturb the logged poses of depth frames and robot pose at every query with uniform noise of up to $\pm0.1$\,m per axis and $\pm0.2$\,rad.

As shown in Table~\ref{tab:noise}, OctoMap + A* using keyframes labeled with exact poses slightly outperforms VNT-PA, but its success rate drops by 42.1 points under localization error. The major reason is that the map commits every frame to the grid at its noisy pose, which mislabels open corridors as occupied. Meanwhile, VNT-PA loses only 8.0 points, where most failures are \textsc{stop} just outside the success radius, because the planner reads its distance to the goal from a perturbed pose. Building the map from all frames mitigates this drop by integrating more observations per voxel, allowing free-space measurements to correct erroneous obstacle observations caused by localization errors. However, VNT-PA still outperforms the conventional pipeline in that case.

\begin{table}[ht]
\centering
\caption{Closed-loop navigation of VNT-PA (single seed) and conventional pipeline with exact and noisy poses}
\label{tab:noise}
\setlength{\tabcolsep}{4pt}
\small
\begin{tabular}{ll cc cc}
\toprule
 & & \multicolumn{2}{c}{Exact poses} & \multicolumn{2}{c}{Noisy poses} \\
\cmidrule(lr){3-4} \cmidrule(lr){5-6}
Context & Method & SR & SPL & SR & SPL \\
\midrule
Keyframes  & OctoMap + A* & \textbf{96.2} & \textbf{92.2} & 54.1 & 46.0 \\
           & VNT-PA       & 94.6 & 91.7 & \textbf{86.6} & \textbf{82.7} \\
\midrule
All frames & OctoMap + A* & 96.1 & 92.1 & 78.1 & 72.4 \\
           & VNT-PA       & \textbf{96.2} & \textbf{93.1} & \textbf{88.0} & \textbf{84.1} \\
\bottomrule
\end{tabular}
\end{table}

\subsection{Summarizer Attention}
\label{sec:summarizer-attention}

\begin{figure}[h]
\centering
\includegraphics[width=1.0\columnwidth]{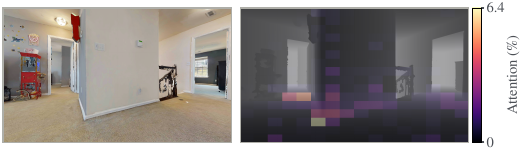}
\includegraphics[width=1.0\columnwidth]{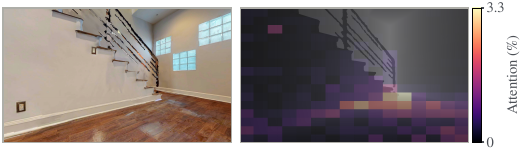}
\caption{Summarizer attention pattern on two sampled frames from evaluation scenes. Left: RGB images, for reference only. Right: attention of the summarizer's learned query token over patch tokens, averaged over attention heads and overlaid on the depth image.}
\label{fig:summarizer-attention}
\end{figure}

To understand what each entry of the spatial context encodes, we visualize the summarizer's attention pattern over patch tokens (Fig.~\ref{fig:summarizer-attention}). The top frame shows a hallway with a wall corner and two doorways. The attention concentrates on the floor right in front of the corner and on the open doorways. The bottom frame faces an open area in front of a staircase, where the attention spreads over the floor and peaks at the bottom of the staircase.

These examples suggest that the summarizer attends to boundaries between free space and obstacles, and also to topologically important regions such as doorways. Both cues are useful for navigation, which requires avoiding nearby obstacles and reasoning about free space connectivity. We hypothesize that the self-attention blocks combine these local descriptions of free space across the environment, so that the goal-aware context acts as an implicit map. The query token, anchored at the robot pose, then reads this implicit map to predict actions toward the goal.

\section{Limitations and Discussion}
\label{sec:limitations}

VNT-PA plans in 2D, consistent with other methods evaluated in the same simulation environment, which assumes a roughly constant camera height during execution. Encoding relative $\mathrm{SE}(3)$ transformations between tokens as in~\cite{gta} provides a natural extension to 3D planning.

Experiments assume that the robot executes discrete actions, with localization error modeled as an independent perturbation at each step rather than as drift that accumulates along the trajectory. Extending the planner to continuous actions in a physics-based simulator and evaluating it under realistic odometry drift are natural next steps.

Finally, we deliberately chose a purely geometric task because it isolates the question of whether pose-stamped experiences can replace explicit maps. However, the architecture operates on arbitrary token embeddings. Extending per-frame features with semantic or language-aligned features would allow a single model to decide both where to go and how to get there from the same retained experience, rather than splitting these roles between a foundation model and a separately maintained map as in [8], [9].

\section{Conclusion}
\label{sec:conclusion}

We propose representing an environment as an unordered set of pose-stamped keyframes, and planning over it with a transformer that uses camera poses as positional embeddings. Our experiments show that this positional embedding provides a beneficial inductive bias. The planner trains faster and handles long and detour-heavy episodes better than models that receive the same keyframes as a temporal sequence, through a recurrent state, or with pose as an input feature. The planner's spatial context can also be extended at test time with frames from other times and trajectories. Moreover, the keyframe filter ties the context size to the space observed rather than to the length of the robot history. Under localization noise, the planner also degrades more gracefully than a conventional pipeline with maps built from the same frames.

Future directions include extending the framework to semantic and language-specified tasks and dynamic scenes, and replacing the encoder with a feed-forward multi-view reconstruction model~\cite{vggt}, whose tokens already encode geometric relationships across views and can also provide camera poses.
\section*{Acknowledgment}
We are grateful to Ariel Barel, Eran Iceland, and Dexter Ong for insightful discussions and feedback that improved this work.

\bibliographystyle{IEEEtran}
\bibliography{refs}

\end{document}